\documentclass[runningheads]{llncs}
\usepackage{algpseudocode,ulem}
\usepackage{algorithm,color}
\usepackage{xcolor}
\usepackage{eucal}

\usepackage{amsmath}
\usepackage{graphicx}

\usepackage{hyperref}
\begin{document}
\title{A Diffeomorphic Aging Model for Adult Human Brain from Cross-Sectional Data}
%
%
\author{Alphin J Thottupattu\inst{1} \and
Jayanthi Sivaswamy\inst{1}\and
Venkateswaran P.Krishnan\inst{2}}
\authorrunning{Thottupattu, Sivaswamy and Krishnan}
%
\institute{International Institute of Information Technology, Hyderabad 500032,
India  \and
TIFR Centre for Applicable Mathematics, Bangalore 560065, India
\\Corresponding Author e-mail: \email{alphinj.thottupattu@research.iiit.ac.in}
}

\maketitle              
\begin{abstract}
 Normative aging trends of the brain can serve as an important reference in the assessment of neurological structural disorders. Such models are typically developed from longitudinal brain image data -- follow-up data of the same subject over different time points. In practice, obtaining such longitudinal data is difficult. We propose a method to develop an aging model for a given population, in the absence of longitudinal data, by using images from different subjects at different time points, the so-called cross sectional data. We define an aging model as a diffeomorphic deformation on a structural template derived from the data and propose a method that develops topology preserving aging model close to natural aging.The proposed model is successfully validated on two public  cross sectional datasets which provide templates constructed from different sets of subjects at different age points.  

\end{abstract}
\section{Introduction}
Human brain morphometry varies with respect to age, gender, and population. 
Since the human brain changes structurally with age, understanding the normative aging process from structural and functional images has been of interest both in general and within a specific population. Studies aimed at arriving at such an understanding, either use a longitudinal or a cross sectional design for collecting images of the study cohort. The former is usually difficult as it is challenging to access a fixed cohort over an extended number of years and scan them repeatedly. A more pragmatic approach is based on a cross sectional design where a set of individuals in different age range forms the cohort. This approach makes it easier to collect scans but their analysis requires a disentangling of the inter-subject variations from age-related changes which is not straight forward. A more elaborate treatment of the differences in such approaches can be found in \cite{aging_decline}.

Regardless of the design, templates play a major role in gaining an understanding of the aging process and derive a normative standard. Templates are images defined using an appropriate reference coordinate space. Templates created for the young adult Caucasian population \cite{icbm,icbm2} are the most well known and used, though population specific templates are also gaining attention \cite{chinese,IBA100}.
In computational anatomy, aging is typically modelled as a \textit{continuous} deformation of a template image over time \cite{diff_growth}. This modelling helps to derive any age-specific template from the model, develop subject specific growth trajectory and derive direct interpretations from the deformation field about the aging pattern.  In this paper, we propose a method to develop such an aging model for the adult human brain from cross sectional data drawn from a specific population. 
Studies using longitudinal data have modelled the aging-based deformation in a variety of ways such as a geodesic \cite{GEOREGRESSION1,GEOREGRESSION2,GEOREGRESSION3} piece-wise geodesic \cite{long_data_peicewise_geodesic}, as a spline \cite{longmodel_spline} as well as with stationary velocity fields parameterized path \cite{base} and acceleration parameterized path \cite{acceleration2,acceleration3}. In a cohort-based longitudinal study, the variability in inter-subject aging trends can also be high. This was handled in \cite{spatio_var1} by considering a tubular neighbourhood for the deformation. The spatio-temporal model suggested in \cite{aging_model} also considers similar variations due to diseased data points in the dataset and uses partial least squares regression to compute normal aging deformations; this gives modes of aging and corresponding scores for each subject.

A cross sectional design allows creation of larger data sets compared to that with longitudinal data. In aging studies with a cross sectional design, the inter-subject variability within an age group and across age groups is disentangled to some extent by developing age-specific templates  \cite{age_spec_atlas2,age_spec_atlas4,age_spec_atlas5}. Several such age-specific templates are publicly accessible
even though the image-sets used for template generation are not publicly available \cite{age_spec_atlas2,age_spec_atlas5,AGE_SEPC_ATLAS1_brains,AGE_SEPC_ATLAS2_Richard}. 

We are aware of only two reports that explicitly develop an aging model from cross-sectional data. In the first \cite{random_pop}, an image regression approach based on weighted averaging is proposed for the aging model. In the second \cite{common_temp_space1}, a global template is derived from all the age-specific templates and the mapping between each of the age-specific template to the global template constitutes the aging model. The consequence of the second approach is that any comparison of a subject image with the global template space needs two transformations of the subject image;  one from the subject image to the corresponding age-specific template and then to the global template image. Further, this is a departure from the notion of aging as a deformation process which acts on a template image \cite{diff_growth}. 

We argue that an aging model represented by a set of image points developed by smoothly deforming a template image is more natural than a weighted average of image points in neighbourhoods. Hence given cross-sectional data, we explore the use of a diffeomorphic deformation of a template image as an aging model. In this paper, we focus on developing the aging model from age-specific templates. The contributions of this paper are: i) a method to derive an aging model from cross sectional data, and ii) an aging model based on a diffeomorphic deformation process applied to the global template for all age points, which is closer to the definition in \cite{diff_growth}. The key features of the proposed aging model are: it aids mapping  of a subject image to the template space with a minimum number of deformations and its ability to handle temporally non-uniform diffeomorphic deformations.

\section{Method}
  
 The proposed method derives an aging model from a given set of cross-sectional data for different age groups. The aging is modelled as a diffeomorphic deformation of a global structural template defined from all the images in the given data.
 The method permits the flexibility to use only age-specific templates instead of the entire dataset as full datasets are usually not publicly available.\\
 The proposed aging model has two elements derived from the supplied templates: (i) a structural template for the brain and  (ii) an aging deformation as a function of time defined on the structural template.
 The diffeomorphic aging, defined as a suitable deformation of the structural template, is explained in Sect. \ref{Comput_G}. Computing the aging deformation, explained in Sect. \ref{Ag}, and mapping it to the  structural template explained in Sect. \ref{mapping}, are the main steps of this method. The model derives temporally and spatially smooth deformations to minimize the effect of cross-sectional variations in the aging deformation computed from the cross-sectional data. These are discussed in detail below.

 \subsection{Computing the structural template}
 Age-specific templates are used as inputs to derive the aging model. Closely and equally spaced age-specific templates, each of them generated from equal number of images is preferred. Let an age-specific template $T_i$ at an interval about age $i$ be a representation of the brain for the given population and assume that we have $N$ such templates. As each of the templates is derived from different sets of images, the template space defined for each set need not be the same and thus the aging path will be different for each $T_i$. Finding the common aging path from the unaligned $T_1, T_{2},\cdots,T_{N}$ is the challenge here. The final aging model is defined using a template $G$ constructed from the set of all $T_i$s in the  diffeomorphic space $\mathcal{G}$. 
 The template $G$ is the best structural representation of all the $T_i$s, which is computed using a non-rigid group-wise registration method called SyNG proposed in \cite{SyNG}. SyNG iteratively computes the templates that minimizes the average distance from the template to each of the $T_i$s on the diffeomorphic space $\mathcal{G}$, and the optimal template is $G$. The distance from this template to each of the $T_i$ incorporates two aspects; the cross-sectional  and aging deformations.

The cross-sectional variations and aging deformations in the data affect $G$ less as it is the template developed from the whole data covering the entire age range of interest. The template $G$ can be considered as the global template for all $T_i$s and considering it as the structural template element in the proposed model avoids biases toward any $T_i$. All the $T_i$s are aligned to $G$ using an affine transformation before developing the model to make it affine invariant. This simplifies the model as an affine alignment can be  done accurately from one template to another or to an image.
 \label{Comput_G}
\\
\subsection{Computing the Aging Deformation}
\label{Ag}
\subsubsection{Assumptions made while computing the aging deformation}
In order to compute the aging deformation from  cross-sectional data, it is useful to understand the natural aging process from a physiological/structural perspective. It is observed that in a mature human brain (from approximately 20 years), brain tissue regions shrink and the ventricular space increases with aging \cite{structural_changes,structural_changes2,doctor}. A key implication of this observation is that the deformation that the brain undergoes with normal aging is spatially and temporally smooth and topology is preserved as no new structure appears with aging. We can therefore assume that growth-induced deformation will be the smoothest and the most predictable among other types of spatial (cross-sectional variation) and temporal (atrophy) deformations. The log-Euclidean framework \cite{symmetric_demon} covers such less-complex diffeomorphic deformations, and can be used to extract aging deformations from cross-sectional data. The space of diffeomorphisms is an infinite-dimensional manifold, and subject-images can be generated by applying a set of diffeomorphisms $\mathcal{G}$ on a template image. The log-Euclidean framework uses the locally Euclidean nature of the manifold to work with diffeomorphisms in a computationally efficient manner. This is defined by representing the diffeomorphism by Stationary Velocity Fields (SVF). Group exponential maps are generally used to compute the deformation $\phi$ represented by the SVF $v$, that is, $\phi=\exp(v)$. \\

\subsubsection{Aging deformation modelled by two SVFs}
Recall that we already have age specific templates $\{T_i\}$ and their global representation $G$. 
The aging deformation is the second element in the aging model. As mentioned in Sect. \ref{Comput_G}, $G$ does not carry any information about the aging deformation. Mapping between $G$ and $T_i$s constitutes both aging and cross-sectional deformations. Therefore $G$ cannot be used directly to extract the aging deformation from the $T_i$s. 

An age-specific template $T_M$ among the $T_i$s that needs the smallest deformation to map $G$ to that template is used as reference template to compute the aging deformation. The aging deformation computed with respect to $T_M$ can be mapped to $G$ fairly accurately as they are close. The SVFs $\{v_i\}$ that maps each $T_i$ to $G$ are computed first to find the reference template $T_M$. Let $\left \| v_i \right \|$ be a measure of the distance between $G$ and $T_i$. Then the desired $T_M$ is the template corresponding to the smallest norm $\lVert v_i\rVert$.
 In other words,    
\begin{equation}\label{TM}
    T_M =  T_i \mbox{ where } i \mbox{  is such that } \lVert v_i\rVert = \mbox{min}\{ \lVert v_k\rVert, 1\leq k\leq n\}. 
   \end{equation} 
The deformation $\exp(v_M)$ which maps $T_M$ to $G$ is used to map the deformation computed with respect to $T_M$ to the common space.

In the proposed model, the aging deformation is considered as a temporal relationship with a consistent trend between subsequent $T_i$s, with $T_M$ being considered as the reference template. An example of consistent trend is the fluid-filled regions in a mature brain increasing in size with aging. This temporally consistent aging deformation is derived from the deformations between template pairs in the forward ($f$) and backward ($b$) directions. For instance, the deformations $v_{j_f}$ between $(T_{M+(j-1)_f},T_{M+j_f})$ for $j=1,2, \cdots, (N-M)$ are the SVF parameterizations for pairs in the forward direction. Similarly, the deformations $v_{j_b}$ obtained by registering template pairs $(T_{M-(j-1)_b},T_{M-j_b})$ for $j=1,2, \cdots, (M-1)$ are the corresponding SVF parameterizations for pairs in the backward direction. The spatial aging trends will be locally consistent and therefore, composing the forward/backward pairwise deformations can be used to extract the consistent trends in the deformation with respect to $T_M$ in both directions. We propose to do this by composing the deformations sequentially using the Baker–Campbell–Hausdorff (BCH) formulation given in Eqn. \ref{BCH} below. This allows compositions of group exponentials to be expressed as a single SVF. Let the velocity vector field obtained as a result of repeated application of BCH formula on the forward (backward) deformations be denoted as $\mathbf{v_{j_f}}$ ($\mathbf{v_{j_b}}$). The vector field $\mathbf{v_{j_f}}$ defines the single SVF parameterization of the forward deformation from  $T_M$ to $T_{(M+j)_f}$ and $\mathbf{v_{j_b}}$ defines the same for the backward deformation from  $T_M$ to $T_{(M-j)_b}$. These velocity fields are computed from Eqn. \ref{seq1} and  Eqn. \ref{seq2} with an initialization of $\mathbf{v_{1_f}}=v_{1_f}$ and $\mathbf{v_{1_b}}=v_{1_b}$.
\begin{equation}\label{seq1}
 \mathbf{v_{j_f}}=\mbox{BCH}(\mbox{BCH}(\cdots (\mbox{BCH}(\mathbf{v_{1_f}},v_{2_f}),v_{3_f}),...),v_{j_f}), j=2...(N-M),
  \end{equation}
\begin{equation}\label{seq2}
\mathbf{v_{j_b}}=\mbox{BCH}(\mbox{BCH}(\cdots (\mbox{BCH}(\mathbf{v_{1_b}},v_{2_b}),v_{3_b}),...),v_{j_b}) , j=2...(M-1).
  \end{equation}
The BCH formula for a pair of 
forward deformations is given in Eqn. \ref{BCH}. Backward deformations can be computed in a similar manner. 
\begin{multline}
\mbox{BCH}(\mathbf{v_{(j-1)_f}},v_{j_f})=\log(\exp(\mathbf{v_{(j-1)_f}})\exp(v_{j_f}))\\=\mathbf{v_{(j-1)_f}}+v_{(j_f )}+\frac{1}{2}([\mathbf{v_{(j-1)_f}},v_{(j_f )}])+\\\frac{1}{12}([\mathbf{v_{(j-1)_f}},[\mathbf{v_{(j-1)_f}},v_{(j_f )}]]+[{v_{j_f}},[{v_{(j_f)}},\mathbf{v_{(j-1)_f}}]])+\cdots =\mathbf{v_{j_f}}.
\label{BCH}
\end{multline}
Here, $[\cdot,\cdot]$ denotes the Lie bracket of two vector fields.\\
It should be noted that since the BCH approximation is valid only for small deformations, in practice, $v_{(j_f)}$ is divided into $n$ smaller deformations such that $\frac{v_{(j_f )}}{n}< 0.5 \times$ voxel dimension, and these smaller deformations are composed iteratively  with $\mathbf{v_{(j-1)_f}}$ to compute $\mathbf{v_{j_f}}$.  In the proposed method, the extracted deformation is constrained to be spatially smooth due to the log-Euclidean framework and temporally smooth since the composing step captures only the temporally consistent trends from the sequential data. For simplicity, the forward aging deformation from $T_M$ to $T_N$,  $\phi_f=\exp(\mathbf{v_{(N-M)_f}}t)$ is denoted as $\exp(\mathbf{v_f}t)$ and the backward aging deformation from $T_M$ to $T_1$,  $\phi_b=\exp(\mathbf{v_{(M-1)_b}}t)$ is denoted as $\exp(\mathbf{v_b}t)$.

The computed aging deformations $\phi_f$ and $\phi_b$ vary uniformly with time which is not consistent with the natural aging trends whereas the aging deformation cannot be expected to vary uniformly, for example tissue degradation will be rapid for elderly age range \cite{doctor}. Hence, a temporal dependency is introduced in $\phi_f$ and $\phi_b$ to accommodate any non-uniform changes in natural aging. This step is explained in the next section.
\label{Comput_v}
\subsubsection{Imposing non-uniform temporal variations on aging deformation}
\begin{figure}
        \centering
        \includegraphics[scale=.7]{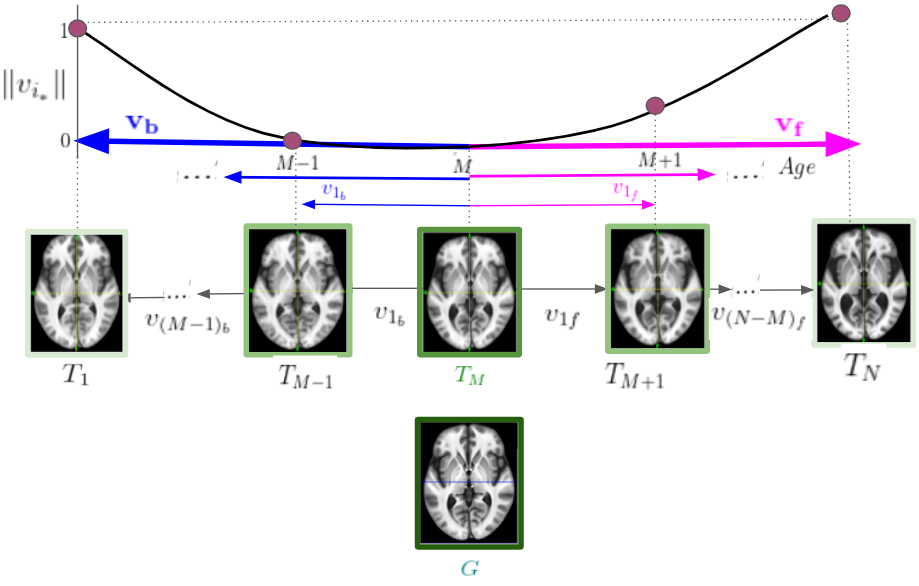}
        \caption{Illustration of the proposed aging model computation framework}
        \label{consis}
\end{figure}

The aging deformation need not  increase linearly in time with respect to $T_M$. Hence we  propose a quantification for the aging deformation (denoted as $R$) at each time point in Eqn. \ref{R}. This is defined in terms of the distance between $T_i$ and $T_{M}$ as in Eqn. \ref{R}. Here $\mathbf{v^*}=\mathbf{v_f}$ in the forward direction and $\mathbf{v_*}=\mathbf{v_b}$ in the backward direction. Further, $\mathbf{ v_i}=\mathbf{ v_{j_f}}$ for $i=(M+j)_f$ and $\mathbf{v_i}=\mathbf{v_{j_b}}$ for $i=(M-j)_b$. With this, let us define
\begin{equation}
    R(i)=\frac{d(T_1,T_i)}{d(T_M,T_i)}=\frac{\left \|\mathbf{ v_i} \right \|}{\left \| \mathbf{v^* }\right \|}.
    \label{R}
\end{equation}
Since $R(i)$ is a discrete sequence, whereas a continuous aging trend is of interest, a smooth curve $\gamma(t)$ is found by fitting a curve to $R(i)$. In our implementation a smooth spline fitting was done in the forward and backward directions. The function $\gamma(t)$ for  $t=[t_0,t_N]$, quantifies the aging deformation at a particular time point with respect to $T_M$. As this deformation increases in both directions with time, the curve will, in general, have a bilateral increasing trend about the age point corresponding to $T_M$. An illustration of the proposed method to extract the aging trends is shown in Fig. \ref{consis}.
\label{temporal_var}
\subsection{Transferring the deformations to the global template space}
\begin{figure}
        \centering
        \includegraphics[scale=.6]{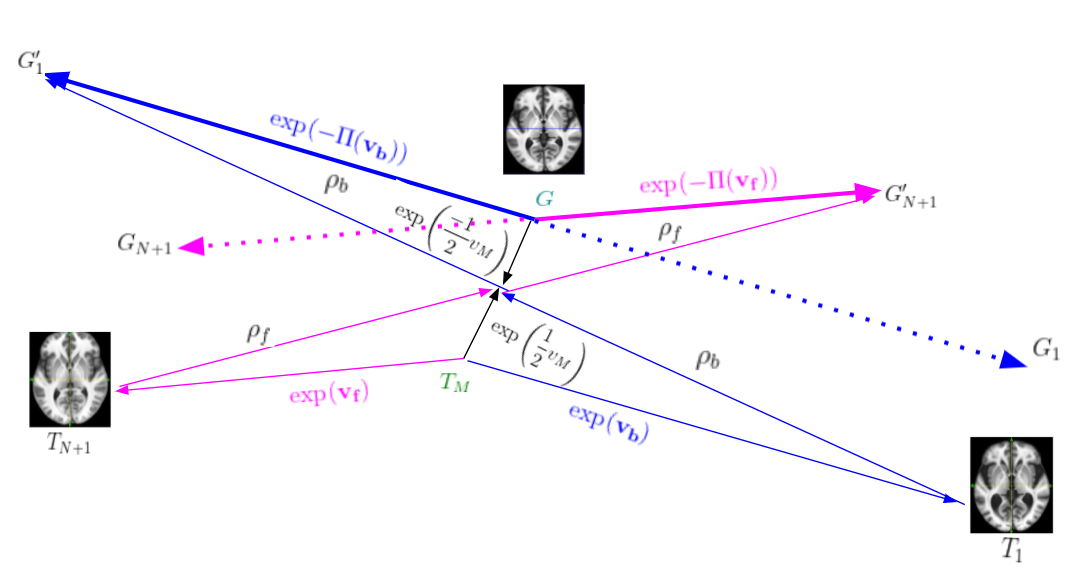}
        \caption{$T_M$ is mapped to $G$ using $\exp(v_M)$ and the path is used to transport $\exp(\mathbf{v_f})$ and $\exp(\mathbf{v_b})$ to the global template space.}
        \label{paralltrn}
\end{figure}

The deformations captured using  Eqn. \ref{BCH} are mapped to the global template space using the mapping from $T_M$ to $G$, i.e., $\exp(v_M)$. The captured deformations  on the manifold $\mathcal{G}$ are parameterized by SVF. In order to transfer the aging deformations to the global template space we use an  existing algorithm \cite{Parelle_Transport} for parallel transport. 
This is explained next.\\
Let the global template space images corresponding to $T_i$s be $G_i$s. The deformations to be transported are parameterized by SVFs $\mathbf{v_f}$ and $\mathbf{v_b}$. A schematic of the deformation mapping scheme is shown in Fig. \ref{paralltrn}.  
Here $\rho_b=\exp\left ( \frac{v_M}{2} \right )
\circ \exp(-\mathbf{v_b})$ and $\exp(\Pi (\mathbf{v_b}))=\exp\left ( \frac{v_M}{2} \right )
\circ\rho_b^{-1}$. Therefore,

\begin{equation}
\exp(\Pi (\mathbf{v_b}))=\exp\left ( \frac{v_M}{2} \right )
\circ \exp(\mathbf{v_b})\exp\left ( \frac{-v_M}{2} \right ),
\label{f}
\end{equation}
and similarly,
\begin{equation}
 \exp( \Pi(\mathbf{v_f}))=\exp\left ( \frac{v_M}{2} \right )
\circ \exp(\mathbf{v_f})\exp\left ( \frac{-v_M}{2} \right ).
\label{b}
\end{equation}

\label{mapping}
 In  Fig. \ref{paralltrn},
 $G_1'= G \circ \exp(-\Pi (\mathbf{v_b}))$ and $G_{N+1}'= G \circ \exp(-\Pi (\mathbf{v_f}))$. Thus, the inverse of the mappings from $G$ to $G_1'$ and  $G_{N+1}'$ i.e., $\exp(\Pi(\mathbf{v_b}))$ and $(\exp(\Pi(\mathbf{v_f}))$ gives $\phi_b$ and $\phi_f$ respectively.
 
\begin{algorithm}[H]
  \caption{Proposed Algorithm}
 \begin{algorithmic}
\State \textbf{Input:} $T_{1},T_{2},\cdots,T_{N+1}$, 
\State \textbf{Result:} Aging Model (${\color{teal} G}$, $ {\color{magenta} \Pi (\mathbf{v_f})}, {\color{blue} \Pi (\mathbf{v_b})}$, $\gamma(t)$)
\State \textbf{Step 1:} Compute the global template as group mean of $T_{1},T_{2},\cdots, T_{N+1}$ $\longrightarrow {\color{teal} G}$
\State \textbf{Step 2:} Register $G$ to $T_i$ $\forall i \in[1...N]$   $\longrightarrow v_{i}$
\State \textbf{Step 3:} Compute the distance $d_i=||(v_i)||$ between $G$ and each $T_i$
\State \textbf{Step 4:} Find $T_i$ which is closest to $G$ by comparing $d_i$ values$\longrightarrow T_M$ 
\\ 

\textbf{Repeat Step 5} and \textbf{Step 6} for $(T_{(M+(j-1))_f},T_{(M+j)_f})$ where $j=1,2...(N-M)$
\State \textbf{Step 5:} Register each pair $(T_{(M+(j-1))_f},T_{(M+j)_f})$ using log-demons registration $\longrightarrow v_{j_f}$
\State  \textbf{Step 6:} Single SVF parameterization of the composed aging deformation from $T_M$ to $T_{(M+j)_f})$  using Eqn.  \ref{seq1} $\longrightarrow \mathbf{v_{j_f}}$ \\
\textbf{Repeat Step 7} and \textbf{Step 8} for $(T_{(M-(j-1))_b},T_{(M-j)_b})$  for $j=1,2,\cdots,(M-1)$
\State \textbf{Step 7:} Register each pair $(T_{(M-(j-1))_b},T_{(M-j)_b})$  using log-demons registration $\longrightarrow{v_{j_b}}$
\State  \textbf{Step 8:} Single SVF parameterization of the composed aging deformation from  $T_M$ to $T_{(M-j)_b}$ using Eqn.  \ref{seq2} $\longrightarrow \mathbf{v_{j_b}}$ 
\State  \textbf{Step 9:} $\mathbf{v_f}\longleftarrow \mathbf{v_{(N-M)_f}}
$ and $\mathbf{v_b}\longleftarrow \mathbf{v_{(M-1)_b}}
$
\State \textbf{Step 10:} Parallel Transport $\mathbf{v_f}$, $\mathbf{v_b}$ along $v_M$ using Eqn.  \ref{f} and \ref{b} respectively $\longrightarrow {\color{magenta} \Pi (\mathbf{v_f})},{\color{blue} \Pi (\mathbf{v_b})}$.
 \State \textbf{Step 11:} Compute a curve fitting for the discrete function R defined by Eqn. \ref{R} using $\mathbf{v_{j_{f}}},\mathbf{v_{j_{b}}},\mathbf{v_f}$ and $\mathbf{v_b}$ $\longrightarrow \gamma(t)$ 
 \end{algorithmic}
 \label{algo}
 \end{algorithm}

\subsection{The aging model}
The aging model has three components, $G$, $\gamma(t)$ and the SVF parameterization of the transported forward and backward deformations $\Pi (\mathbf{v_f}),\Pi (\mathbf{v_b})$  respectively. An age-specific template at any time point $t$ can be computed using the following formula: 
\begin{equation}
T(t)=\begin{cases}
 G \circ \exp(\Pi (\mathbf{v_f})\gamma(t)) \mbox{ for } t\geq M, \\ 
  G \circ \exp(\Pi (\mathbf{v_b})\gamma(t)) \mbox{ for } t\leq M. 
\end{cases}
\label{Ttb}
\end{equation}

\section{Experiments}
 In this section, We report on valiadtion of the proposed model and experiements with the model. All experiments, barring the one with simulated data, were done on 3D data though only 2D central slices from the results are shown for visual comparison. The proposed aging model is affine invariant, and therefore results  were also aligned using affine transformation prior to comparison. We have used two databases to do experiments which are discussed in Sect. \ref{data} and the models derived from these datasets are discussed in Sect. \ref{model}. Experiments done to i) assess the quality of representation of the proposed model are presented in Sect. \ref{sub1}  and ii) validate the model are presented in Sect. \ref{sub2}. 
 
\subsection{Data}
 The proposed method to create an aging model was implemented using two cross-sectional template datasets: (i)Brain Imaging of Normal Subjects (BRAINS) \cite{AGE_SEPC_ATLAS1_brains} and (ii)Neurodevelopmental MRI Database \cite{AGE_SEPC_ATLAS2_Richard} (NEURODEV). Only age-specific templates are accessible in these datasets along with information on the age interval and number of scans of subjects used to create each age-specific template. From Table \ref{table1}, it can be observed that in BRAINS, the inter-age interval is not uniform, particularly at the upper age level, and the number of scans used for template creation is less relative to NEURODEV.  The inter-age interval is shorter (5 years) and uniform in NEURODEV.
 \begin{table}[]
\centering
\begin{tabular}{|c|c|c|c|}
\hline
\multicolumn{2}{|c|}{\textbf{BRAINS}\cite{AGE_SEPC_ATLAS1_brains}}                                                                                                                            & \multicolumn{2}{c|}{\textbf{NEURODEV}\cite{AGE_SEPC_ATLAS2_Richard}}                                                                                                                             \\ \hline
\begin{tabular}[c]{@{}c@{}}Age interval\\ (Template age)\end{tabular} & \begin{tabular}[c]{@{}c@{}}Number of scans \\ used for\\  template creation\end{tabular} & \begin{tabular}[c]{@{}c@{}}Age interval\\ (Template age)\end{tabular} & \begin{tabular}[c]{@{}c@{}}Number of scans \\ used for\\  template creation\end{tabular} \\ \hline
25-34 (29.5)                                                          & 20                                                                                       & 20-24  (22)                                                           & 559                                                                                      \\ \hline
35-44 (39.5)                                                          & 24                                                                                       & 25-29  (27)                                                           & 525                                                                                      \\ \hline
45-54 (49.5)                                                          & 23                                                                                       & 30-34  (32)                                                           & 422                                                                                      \\ \hline
55-64 (59.5)                                                          & 13                                                                                       & 35-39  (37)                                                           & 73                                                                                       \\ \hline
71-74  (72.5)                                                         & 47                                                                                       & 40-44  (42)                                                           & 96                                                                                       \\ \hline
75-78 (76.5)                                                          & 50                                                                                       & 45-49  (47)                                                           & 82                                                                                       \\ \hline
91-93 (92)                                                            & 48                                                                                       & 50-54  (52)                                                           & 72                                                                                       \\ \hline
                                                                      &                                                                                          & 55-59  (57)                                                           & 81                                                                                       \\ \hline
                                                                      &                                                                                          & 60-64 (62)                                                            & 57                                                                                       \\ \hline
                                                                      &                                                                                          & 65-69 (67)                                                            & 71                                                                                       \\ \hline
                                                                      &                                                                                          & 70-74 (72)                                                            & 65                                                                                       \\ \hline
                                                                      &                                                                                          & 75-79 (77)                                                            & 44                                                                                       \\ \hline
                                                                      &                                                                                          & 80-84 (82)                                                            & 44                                                                                       \\ \hline
                                                                      &                                                                                          & 85-89 (87)                                                            & 20                                                                                       \\ \hline
\end{tabular}
\caption{Composition of the two age-specific template datasets \cite{AGE_SEPC_ATLAS1_brains,AGE_SEPC_ATLAS2_Richard} used in our study.}
\label{table1}
\end{table}
\label{data}

\subsection{Aging Model}
\begin{figure*}        
     \centering
     \includegraphics[scale=.5]{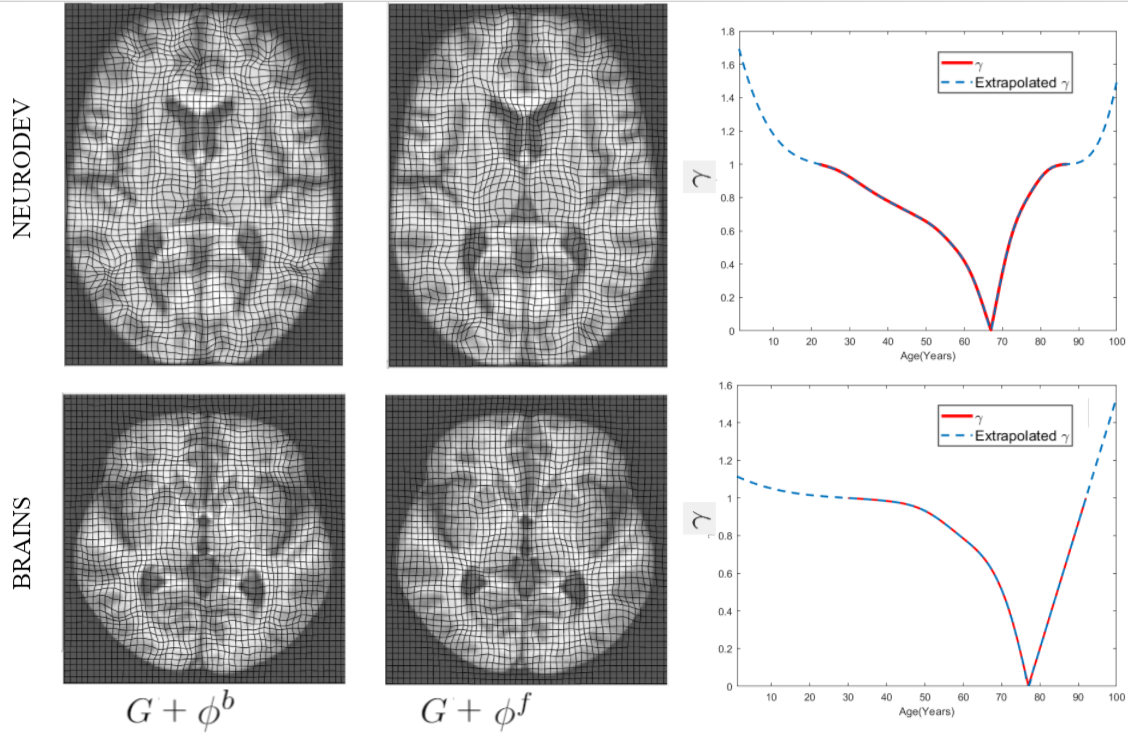}
     \caption{The aging model computed with NEURODEV and BRAINS datasets}
     \label{ag_result}
 \end{figure*}

Recall that the proposed aging model has two elements, namely, the structural template $G$ and the aging deformation. The aging deformation has three components: the forward aging deformation $\phi_f$ parameterized by $\mathbf{v_f}$, $\phi_b$ parameterized by $\mathbf{v_b}$ and the $\gamma$ function. The proposed model developed  with NEURODEV and BRAINS datasets are shown in Fig. \ref{ag_result}. A direct interpretation of $\gamma(t)$ plot does not give much information about the aging trend as it represents the degree of deformation with respect to $G$, rather than any of the end point templates. It however does indicate the age point that corresponds to the reference template $T_M$. In the case of NEURODEV this is 67 years and for BRAINS it is 77 years where the template age ranges for these datasets are 22-87 and 30-92 years respectively.
 
  \label{model}
\subsection{Representation Quality Analysis}
\label{sub1}
Age-specific templates were generated with the proposed aging model using Eqn. \ref{Ttb}, and were used for visual comparison to assess the quality of representation. Comparisons are done with natural aging trends in Sect. \ref{sub1a}, existing spatio-temporal atlas \cite{aging_model} in Sect. \ref{sub1b} and the supplied templates used for model creation in Sect. \ref{sub1c}.
  
\subsubsection{Compatibility with Natural Aging}
Templates at increasing age points were generated with the proposed aging model to study the structural change with aging. The BRAINS dataset \cite{AGE_SEPC_ATLAS1_brains} was chosen to do this experiment as it covers a longer span at the elderly age end where more changes are expected. Human brain aging literature \cite{age1,age2,doctor} indicates that a mature brain undergoes minimal cognitive and structural changes up to the age of $\approx50$ and more for the elderly, i.e. $\approx 60+$. This trend was verified by computing the intensity difference between the current template and the first (at age 30) template. This difference essentially is due to age-induced structural change.
\label{sub1a}
\subsubsection{Growth Trend across Aging Models}
Huizinga et al. in \cite{aging_model} proposed a cross-sectional spatio-temporal reference model for representing aging. This model does not ensure a diffeomorphic aging deformation and the template space representation of a subject image needs a computationally intensive group-wise registration with a training set used to generate the model. In contrast, our model requires only one pairwise registration from a subject to the corresponding age-specific template, derived from the model. The aging trends observable in the templates derived from our model was was compared with those derived using \cite{aging_model}; the latter templates are available in \url{http://www.agingbrain.nl/} for the age range of 45-92 years. Templates at the same age points were generated with the proposed method using the NEURODEV dataset. 

 \label{sub1b}

\subsubsection{Age-specific Template Assessment}
The generated templates with our model were visually compared with the templates given in the 2 datasets to understand how well the model have represented these templates. The templates for the first and last time points in our aging model have undergone maximum deformation compared to those at other age points. Hence, such a visual comparison is of interest. 
\label{sub1c}

\subsection{Aging Model Validation}
\label{sub2}
Model validation was done by analysing the ability of the model to capture natural deformations in Sect. \ref{section2a}; and the similarity of model-generated age-specific templates to a set of subject images of same age in Sect. \ref{section2b}. Since our model was derived for a cross-sectional setting, we also studied its performance in a longitudinal data setting as it is of interest. This is presented in Sect. \ref{section2c}.

\subsubsection{Topology Preservation}
Since diffeomorphic deformations best fit \textit{natural} deformations, we considered aging related deformation also as a diffeomorphism. Accordingly, our model is defined on a manifold $\mathcal{G}$ of diffeomorphisms. It is of interest to verify if an extrapolation of the model generates deformations in $\mathcal{G}$ itself. This was done by extrapolating the aging trend and deriving templates in both younger and older ages.  NEURODEV data  which covers that age range of 22-87 (reference template age point, M=77 years) is used for this experiment. The templates from extrapolation in both directions were generated for this experiment with Eqn. \ref{Ttb}.
\label{section2a}

\subsubsection{Validation with Segmentation}
A localised assessment, i.e., of few structures, is of interest in many situations. This requires labeling by aligning the subject image to a labeled template and doing a label transfer. An alignment process that requires smaller deformations indicates that the template is  structurally very close to the subject image. This will lead to better segmentation.  With our age model, this involves only a single registration step and hence potentially least deformation. 
This is in contrast to the steps required when using the model in \cite{common_temp_space1} which requires two registration steps: one to transfer the labels from the global template to the age-specific template that is closest to the given subject age, and a second to transfer the age-specific template labels to the subject image. Each of these registration steps can contribute to error in labeling in addition to the structural dissimilarity of template and subject image. 

An experiment was done to quantitatively compare the accuracy of labeling using the proposed method and with \cite{common_temp_space1}. Ten subject images from MICCAI 2012 dataset \cite{MICCAI} of ages 18,30,38,45,54,61,68,75,83 and 90 along with the ground truth labels were used to perform the comparison. The templates corresponding to the subject ages 18 and 90, were constructed by  extrapolating the proposed aging model. The accuracy of label transfer from a template is highly influenced by the registration method and the global template labels being used. Hence, efforts were taken to do a fair comparison as follows.

The Neurodev \cite{AGE_SEPC_ATLAS2_Richard} templates were used to study both the aging models. A global template $G$ was constructed with ANTs template creation algorithm \cite{ants}. Starting with a common labeled template \cite{IBA100}, the labels were transferred to the global template using DRAMMS non-rigid registration method \cite{dramms}. The registration algorithms and the parameter setting were fixed to be the same for both aging models. The Dice score was used for assessing the segmentation accuracy. 
\label{section2b}

\subsubsection{Validation with Simulated Longitudinal Data}
The proposed method was aimed at handling cross-sectional data. In order to understand how the model would handle longitudinal data, an experiment was done using simulations as longitudinal data is unavailable. Simulated data was generated as follows. A set ($S$) of fifty randomly deformed Shepp-Logan phantoms were taken (to simulate a cohort) and five copies were made. Deformations with increasing degree was applied on these five copies to simulate aging of different subjects. The five sets thus form our longitudinal data. For each of the five sets a template was computed separately using the method suggested in \cite{ants}. The templates were then used as inputs for the proposed model to generate templates at different age points. These were then compared against the deformed versions of the Shepp-Logan phantom (proxy ground truth).
\label{section2c}

\section{Results}
    
\subsection{Estimation of the Aging Model}

\subsubsection{Compatibility with Natural Aging}
\begin{figure*}         
     \centering
     \includegraphics[scale=.5]{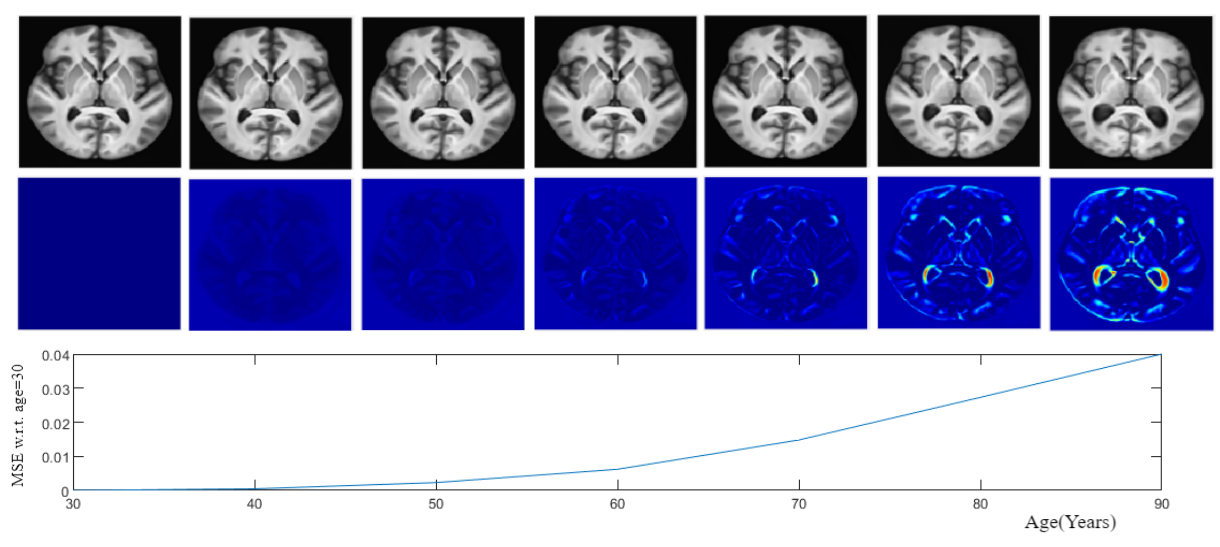}
     \caption{ The proposed aging model at different time points(first row) along with the difference image with respect to the initial time point(second row)}
     \label{result_sim1}
 \end{figure*}

Fig. \ref{result_sim1} shows the generated sequential templates (first row) and difference between the sequential templates and the first template (second row). The difference images facilitate understanding the structural changes with aging. The mean squared error or MSE (difference) is plotted in the third row. The difference appears to be very low for the first few decades relative to the last few decades where changes like ventricular expansion occurs. This trend is consistent with the existing information about natural healthy aging.
 
\subsubsection{Growth Trend across Aging Models}
\begin{figure*}
     \centering
     \includegraphics[scale=.9]{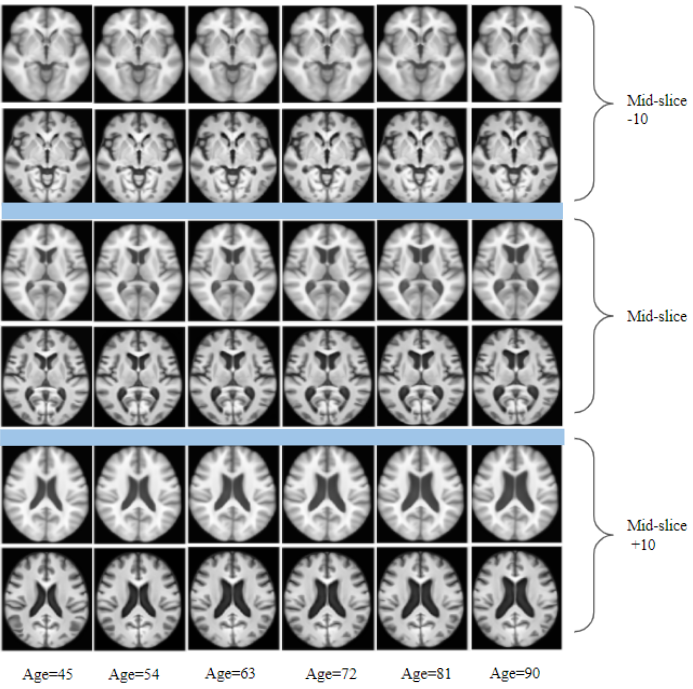}
     \caption{Correctness of Aging trends captured in the model: Publicly available spatio-temporal images \cite{aging_model}(row 1,3,5) compared with images generated at same time points with proposed aging model using NEURODEV data at different time points(row 2,4,6) Each highlighted row pairs compare same slices as specified in the figure.}
     \label{result_sim2}
 \end{figure*}

 Fig. \ref{result_sim2}, shows sample 2D slices of templates from  \cite{aging_model} in odd numbered rows, along with the ones derived with the proposed model (from the NEURODEV dataset) in even numbered rows, for comparison. The comparison at image-level comparison is not meaningful as the templates are generated from different data-sets. However, one can observe growth trends. The structural similarity across rows in a column appear to have similar trends across age indicating growth trend to be consistent. 
\subsection{Age-specific Template Assessment}
\begin{figure*}
     \centering
     \includegraphics[scale=.6]{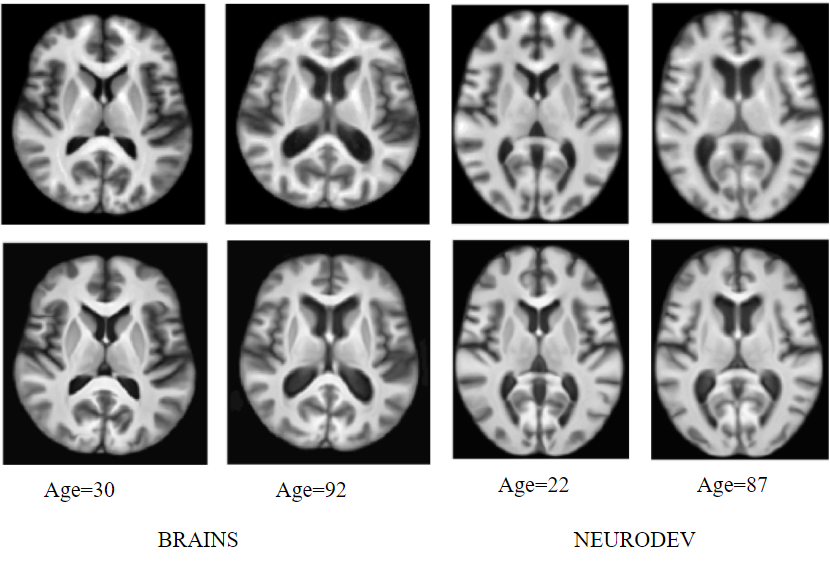}
     \caption{ Comparison of templates given in the two datasets (BRAINS and NEURODEV) (top row) with those generated by the proposed aging model (bottom row). Only the templates for the first and last time-points are shown.}
     \label{result1}
 \end{figure*}

 The given templates along with our generated templates are shown in Fig. \ref{result1} for comparison. The first and last time points for the both BRAINS and NEURODEV are shown in the top row, while the corresponding templates generated by the proposed model are shown in the second row. As per the proposed aging model, the template for the first and last time points are maximally deformed with respect to the template closest to $G$, i.e., $T_M$. Yet, the derived templates are visually quite similar to the templates from the two datasets. Thus, the proposed model appears to preserve the structural details of the given template at each time point.
\subsection{Aging Model Validation}

\subsubsection{Topology Preservation}
\begin{figure*}
     \centering
     \includegraphics[scale=.7]{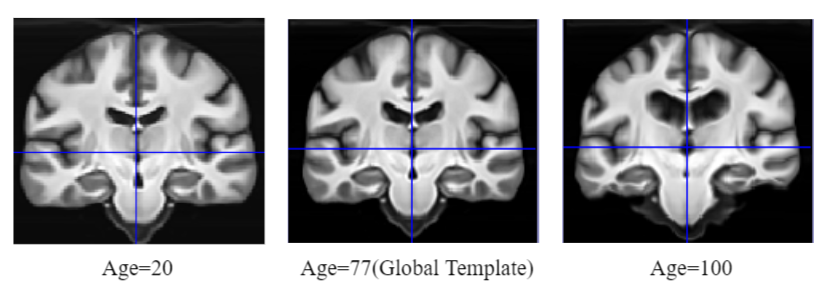}
     \caption{The central coronal slices of extrapolated age-templates are shown along with the global template image}
     \label{extrapolation}
 \end{figure*}

 Two templates, namely at age 20 and age 100, generated with the proposed model are shown in Fig. \ref{extrapolation} along with the Global template. These are results of extrapolation from the data given in the NEURODEV dataset\cite{AGE_SEPC_ATLAS2_Richard}. The topology appears to be preserved even when the aging model is extrapolated in both directions implying that the extrapolated deformations also belong to $\mathcal{G}$. It can also been that while global similarity (in structure) exists across age, local deformations persist. For instance, the ventricle is much smaller at age 20 and enlarges with age, consistent with the expected aging trend. \\
\subsubsection{Validation with Segmentation}
Table \ref{table_dice} lists the Dice scores for segmenting 4 pairs of sub-cortical structures with our model and \cite{common_temp_space1}. It can be observed that the proposed model outperforms \cite{common_temp_space1} as it has higher average (over 10 subjects) dice score. The proposed model's better performance in label transfer implies that it generates most accurate age-point templates which is closest to each given subject image age-point.
\begin{table}[]
\begin{tabular}{|l|l|l|l|l|l|l|l|l|}
\hline
                                                          & \begin{tabular}[c]{@{}l@{}}Right\\ Hippo-\\ campus\end{tabular} & \begin{tabular}[c]{@{}l@{}}Left\\ Hippo-\\ campus\end{tabular} & \begin{tabular}[c]{@{}l@{}}Right\\ Pallidum\end{tabular} & \begin{tabular}[c]{@{}l@{}}Left\\ Pallidum\end{tabular} & \begin{tabular}[c]{@{}l@{}}Right\\ Putamen\end{tabular} & \begin{tabular}[c]{@{}l@{}}Left\\ Putamen\end{tabular} & \begin{tabular}[c]{@{}l@{}}Right\\ Thalamus\end{tabular} & \begin{tabular}[c]{@{}l@{}}Left\\ Thalamus\end{tabular} \\ \hline
\cite{common_temp_space1}                                     & 0.6266                                                          & 0.6229                                                         & 0.6604                                                   & 0.5803                                                  & 0.6802                                                   & 0.6444                                                  & 0.6944                                                   & 0.6917                                                  \\ \hline
\begin{tabular}[c]{@{}l@{}}Proposed \\ model\end{tabular} & \textbf{0.6724}                                                 & \textbf{0.6900}                                                & \textbf{0.7720}                                          & \textbf{0.7266}                                         & \textbf{0.7798}                                          & \textbf{0.7880}                                         & \textbf{0.7912}                                          & \textbf{0.7682}                                         \\ \hline
\end{tabular}
\caption{Model validation via segmentation. Average Dice scores computed over 10 subject images are listed.}
\label{table_dice}
\end{table}
\subsubsection{Validation with Simulated Longitudinal Data}
\begin{figure}
 \centering
     \includegraphics[scale=0.5]{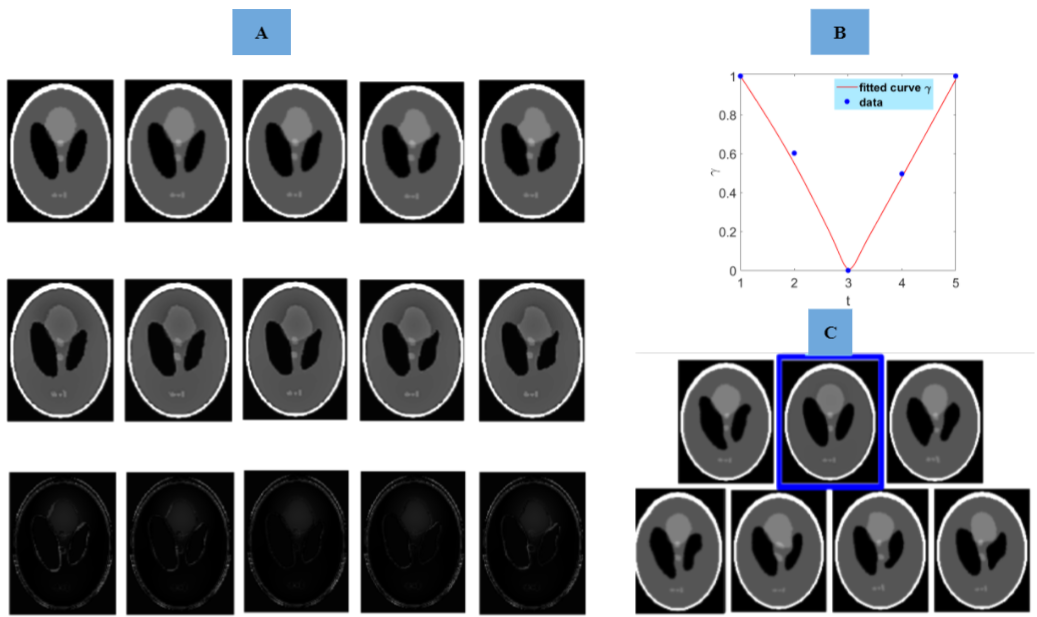}
     \caption{ Aging model for a simulated longitudinal dataset. A - First row: Deformed images with known transformation and second row: images generated with the proposed model, for the same time points (as in the first row)  and last row: the absolute difference of first and second rows; B - The $\gamma$ curve of our aging model and C - Some sample images used in $S$}
     \label{sim}
 \end{figure}

 Sample images generated by applying the simulated deformations on the Shepp-Logan phantom are shown in the first row of Fig. \ref{sim}. This forms the ground truth. The template images derived with the proposed model are shown in the second row. The images in the 2 rows appear to be very similar to each other at the same time points. The template images generated with the proposed model and corresponding $\gamma$ curve is also shown in the same figure. The degree of deformation in the simulated deformation is uniformly increasing with time and hence it can be expected that the $\gamma$ curve will be symmetric with respect to mid-time point. We see that, in Fig. \ref{sim}, is indeed true. The proposed model captures the applied deformation without much errors from the simulated longitudinal data. 

\section{Discussion}
Cross-sectional images at different age points are easier to acquire than that of the same subject. This motivated us to develop a method to generate an aging model using cross-sectional data. The aging model is based on continuous deformation applied to a template. Experimental results show that our aging model can be used to generate templates at different time points in a manner that is consistent with the natural aging trend observed by other studies; it preserves structural details of the supplied templates and generates topology-preserving aging deformations.
\subsection{Limitations}
The proposed aging model has a few  limitations. Firstly, it is applicable only for matured brain growth where no new brain structures are introduced. The quality of the proposed model is completely dependent on the data. Consequently, the number of scans in each age interval needs to be large to generate the age-specific templates that are representative of the cohort/population under study. What constitutes 'large' is an open question. Secondly, while the model reduces the effect of cross-sectional data induced variations in the aging deformation, there is no formal proof as yet that it completely removes the cross-sectional variation. Finally, the proposed aging model defines a single average growth path and does not attempt to model the cross-sectional aging variations. 
\subsection{Future Work}
Since the proposed model works only to obtain a mean aging path, future work can be a refinement in terms of defining the aging model as a distribution of paths about the average path. The basic requirement to develop such a model however, is the availability of scans at different age points, not the templates alone. Our current work is directed at developing a public database for this purpose with subject scans at different age points. 

 \section{Conclusion}
 We have proposed a novel aging model from cross-sectional data. The spatio-temporal smoothness and consistency are assured in the model to make it closer to natural aging. The model has the potential to be used for clinical purposes. Currently population specific aging trends are of interest and this can be generated with the proposed model with less efforts. The code to generate proposed the aging model has made publicly available in \url{http://dx.doi.org/10.17632/nw983x225c.1}. 

\end{document}